# Dynamic behavior analysis for a six axis industrial machining robot

BISU Claudiu[1,a], CHERIF Mehdi[2,b], GERARD Alain [1,c] and K'NEVEZ Jean-Yves[2,d]

[1]University Politehnica of Bucharest, Department of Machines and Systems of Production, 313 Spl. Independentei, sect.6, 060042 Bucharest.

[2]University Bordeaux 1, -I2M- Material, Processes, Interactions-CNRS, UMR 5295, 351 cours de la Libération, 33405 Talence, France.

[a]claudiu.bisu@upb.ro , [b]mehdi.cherif@iut.u-bordeaux1.fr , [c]ajr.gerard@gmail.com , [d]jean-yves.knevez@u-bordeaux1.fr

**Abstract**
The six axis robots are widely used in automotive industry for their good repeatability (as defined in the ISO92983) (painting, welding, mastic deposition, handling etc.). In the aerospace industry, robot starts to be used for complex applications such as drilling, riveting, fiber placement, NDT, etc. Given the positioning performance of serial robots, precision applications require usually external measurement device with complexes calibration procedure in order to reach the precision needed. New applications in the machining field of composite material (aerospace, naval, or wind turbine for example) intend to use off line programming of serial robot without the use of calibration or external measurement device. For those applications, the position, orientation and path trajectory precision of the tool center point of the robot are needed to generate the machining operation. This article presents the different conditions that currently limit the development of robots in robotic machining applications. We analyze the dynamical behavior of a robot KUKA KR240-2 (located at the University of Bordeaux 1) equipped with a HSM Spindle (42000 rpm, 18kW). This analysis is done in three stages. The first step is determining the self-excited frequencies of the robot structure for three different configurations of work. The second phase aims to analyze the dynamical vibration of the structure as the spindle is activated without cutting. The third stage consists of vibration analysis during a milling operation.

## 1 Introduction

The evolution of the performance of robots and programming software provides new machining solutions. For complex parts, six axes robots offer more accessibility than a machining center CNC 5 axis and allow the integration of additional axes to extend the workspace. Robots have seen in recent years an expansion of their field of use with new requirements related to the increasing use of composites. The robots are then considered for machining operations (polishing, cutting, drilling etc.) that require high performance in terms of position, orientation, followed by trajectory precision and stiffness [1], [2], [3], [4]. For drilling operations, the performance of position and orientation of the Tool Center Point are high priority. During the off-line programming of robots machinists, many factors are degrading the accuracy of the machining operation performed [5]. As part of the proposed study, we focus on dynamical phenomena associated with the power chain transmission of a poly-articulated industrial robot KUKA KR240 6-axis-2.

The objective of this work is to characterize the dynamical behavior of the robot to point out the influence of the task position in the robot workspace concerning the dynamical response of the structure. This analysis is done in three stages: the first step is determining the self-excited frequencies of the robot structure in different configurations of work. The second step aims to analyze the dynamical vibration of the robot structure. During this step, the spindle is activated but without cutting in orders to highlight the impact frequencies in the dynamic case. In the last step, measurement and analysis of the robot structure vibration are conducted during a milling test.

## 2 Experimental setup

To achieve this research an experimental device is designed to obtain the dynamical information provided by the system robot/tool/workpiece. The experiments were performed on a 6 axis robot centre (figure 1) with 18 kW of power of the spindle motor and a maximum rotation speed of 42000 rpm. During the experiment, the recording data of vibrations, cutting forces signals in the same time with rotational speed is absolutely necessary. A three-dimensional PCB piezoelectric accelerometer fixed on the robot's head; a National Instruments NI USB-4432 analogical/digital board and Fastview software were used for vibrations measurement. The speed of rotation is achieved through a laser sensor tachometer fixed on the spindle holder. The three-linear directions (X, Y, Z) of the robot corresponding with the three-dimensional axis of the accelerometer. For the cutting forces measurement, a 6D forces dynamometer is positioned between the end of the robot and the HSM spindle (figure 1).

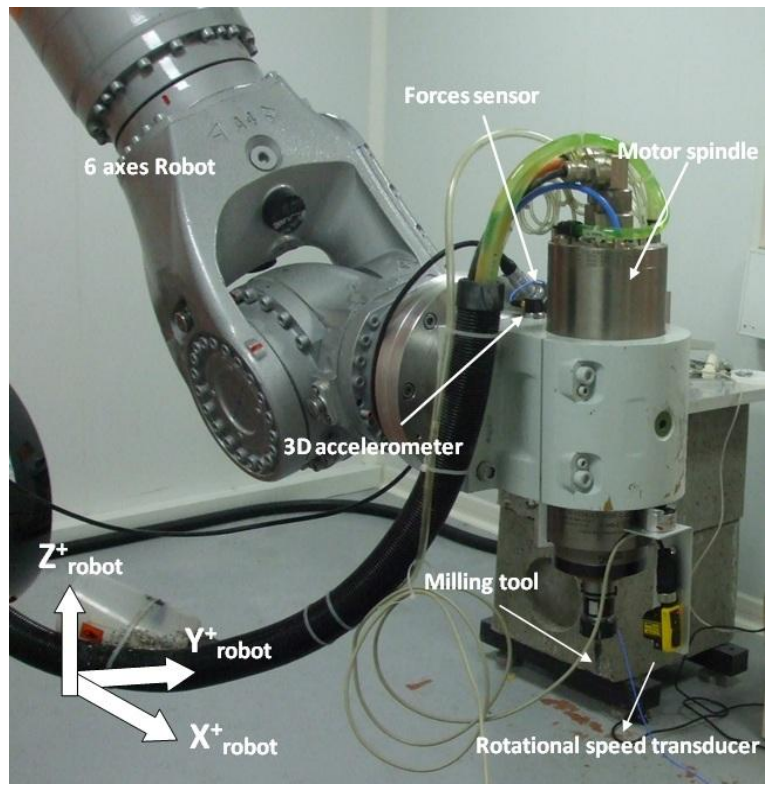

Figure 1: Experimental device.

## 3 Frequency impact analyses

To achieve the objective of this work, a detailed study is carried out on the impact frequencies in three configurations of the robot position. The robot's head was also analyzed by impact vibrations using a PCB piezoelectric three-dimensional accelerometer and an instrumented hammer with a PCB force transducer in order to identify their transfer function in a broad range of frequencies. Samples were recorded at 25 kHz. Figure 2 shows the experimental device where the signal tests is achieved by a National Instruments data board NI USB 4432 and Fastview data processing. The impact tests were performed in three different positions as presented in figure 3. For those configurations, the self-excited frequency on X, Y and Z direction during the impact for each direction are analyzed. The behavior of the cutting process can be modified by the varying self-excited frequency of the robot/tool and workpiece, causing vibrations during the cutting process.

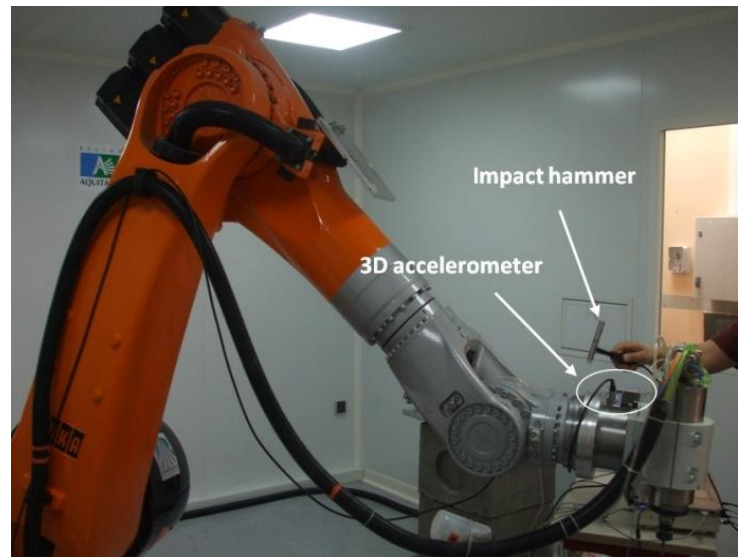

Figure 2: Experimental device for frequency impact.

These vibrations may reflect the dynamic phenomena in the cutting process that is very detrimental to the machined surface, the tool life and even the spindle. [6], [7] [8]. Impact tests take place in three configurations defined in the experimental protocol, representing some of the common positions in cutting composite parts. Following this analysis as follows: $P_1$ configuration, the robot arm is in a position closest to the robot base, followed by position $P_2$ and $P_3$ which crossed the robot moves on a well defined trajectory to the point $P_1$, presented in figure 3.

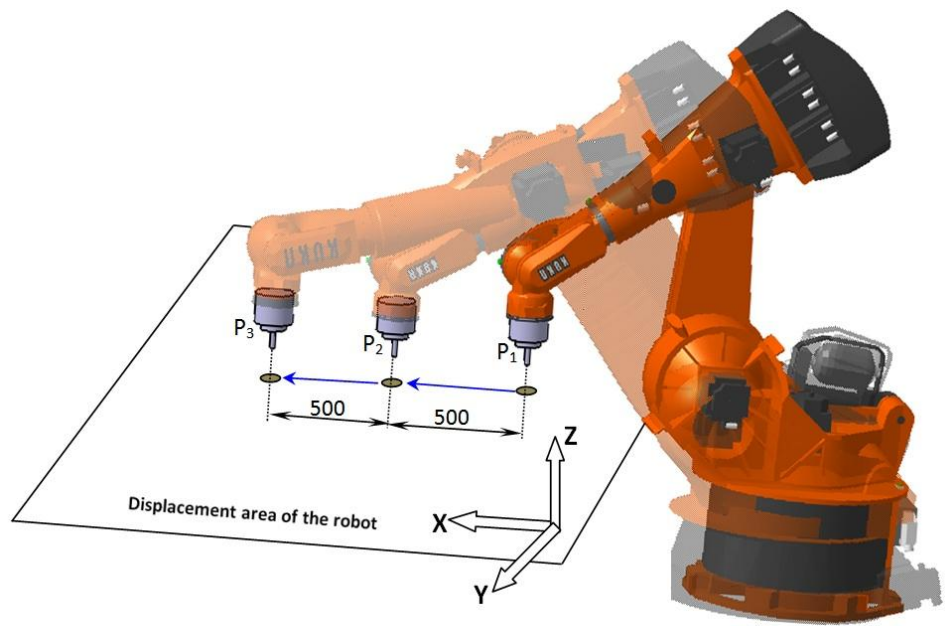

Figure 3: The experimental position for impact test.

The results of the frequency spectra measurements for each configuration (Figure 3) $P_1$, $P_2$, and $P_3$ are presented on figure 4. It can be observed the two main frequency ranges that we will call: LFR- Low Frequency Range, with a frequency range between: 0-250 Hz and HFR- High Frequency Range, with a frequency range between: 1200-3600 Hz. The importance of this study is to identify the robot stiffness, considering that it presents low frequency [9] it has to determine these frequencies in different configurations

7-10 September 2011 – Bucharest – Romania                                                                                    ICASAAM 2011

in order to optimize de cutting robot parameters in order to reduce the effect of the low frequencies respectively the low stiffness of the robot.

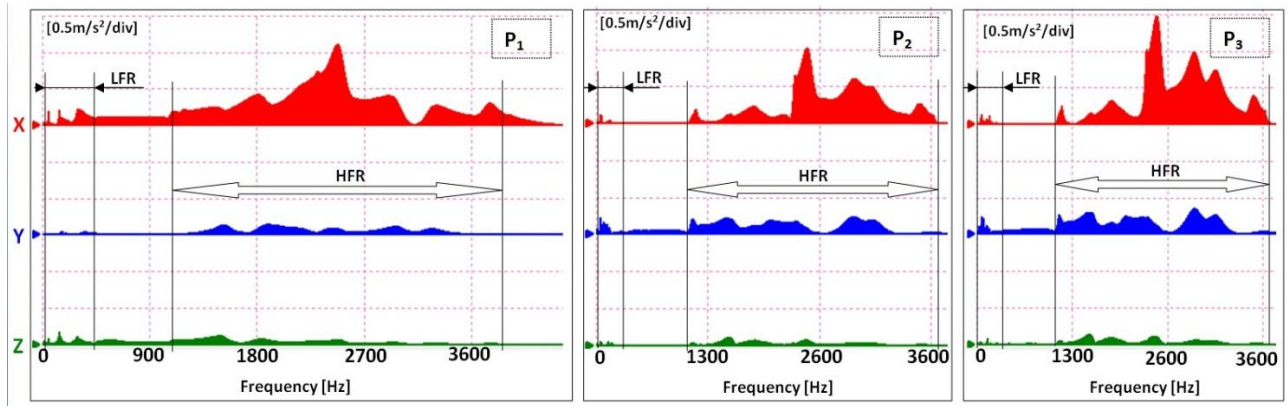

Figure 4: Frequency spectrum for $P_1$, $P_2$ and $P_3$ position

Given the objective characterization of rigidity for various working positions of the robot, we are interested in low frequency range, where the basic rigidity of the robot is emphasized [9]. In order to identify this basic rigidity, a low frequency analysis is made of the frequencies measured during the impact tests in the range 0-250 Hz. The acquisition data is made by continuous acquisition with a sample rate of 6250 samples / sec,) a buffer size of 32 768 samples, and block size of 2000 samples.

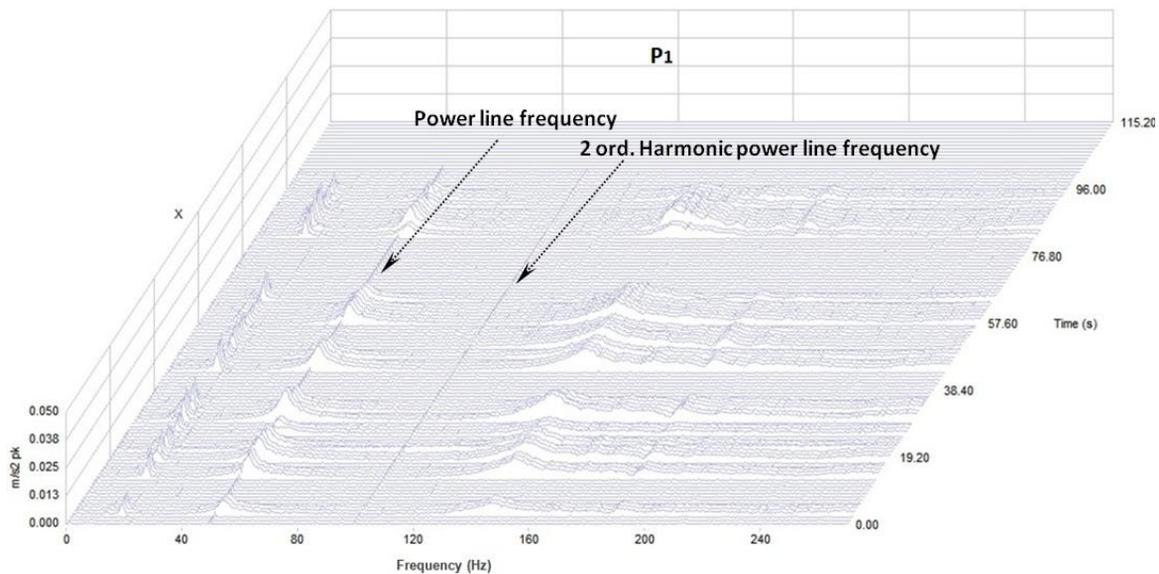

Figure 5: Waterfall frequency diagram for X direction in $P_1$ position

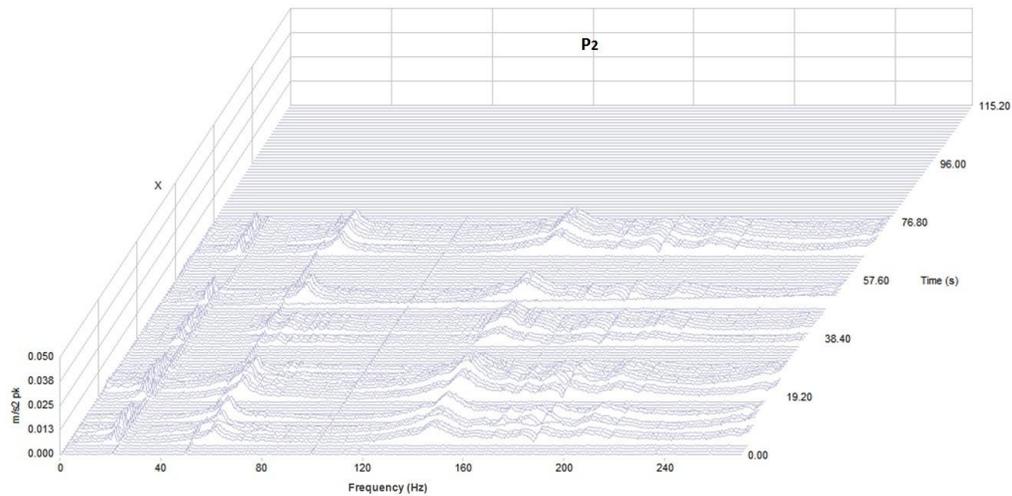

Figure 6: Waterfall frequency diagram for X direction in $P_2$ position

Based on signal processing, Waterfall diagram are used to identify the self-excited frequencies generated by hammer impact excitation. Waterfall diagram is necessary in this analysis because the impact hammer tests are made on-time and we can highlight self-excited frequencies and also their dissociation in relation to external frequencies. Thus the quality of the analysis can easily identify the electrical grid frequency of 50 Hz and the 100 Hz. For the X measurement direction from the impact excitation, the spectrum is observed similar for every tested position. But a slight shift of the fundamental lower frequency (7-10 Hz) to the increase frequency is observed. The $P_1$ position presents the value of 17 Hz (figure 5), while frequency increases to $P_2$ with 20 Hz (figure 6) and 22 Hz for $P_3$ (figure 7). Increasing the frequency variation at position $P_1$ to position $P_2$ and $P_3$ in X direction shows an increase in stiffness.

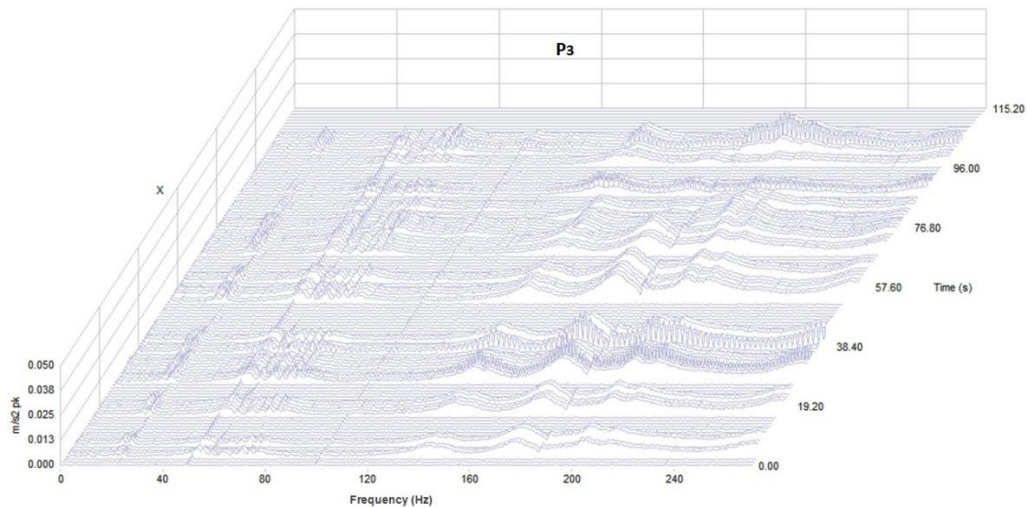

Figure 7: Waterfall frequency diagram for X direction in $P_3$ position

The same situation existed for 50.5 Hz frequency, which $P_1$ position is similar to the frequency of the electric network. This frequency increases reaching 53 Hz for $P_2$ and 55Hz for $P_3$. The same evolution is finder in the case of 130-230 Hz frequency range. This observation shows that the X direction (the axial direction), the stiffness presents an increase when the TCP (tool center point) of robot go through the different position $P_1$, $P_2$ and $P_3$.

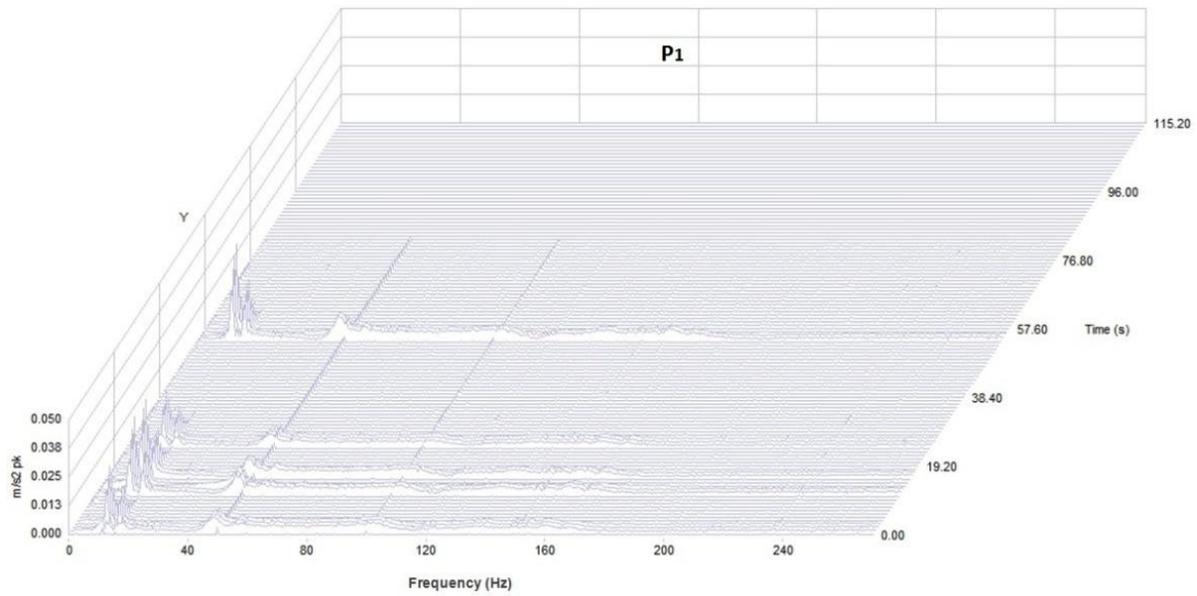

Figure 8: Waterfall frequency diagram for Y direction in $P_1$ position

For the Y direction (Figure 1) frequencies decrease slightly as the robot arm goes away from the position $P_1$, (figure 8). Frequencies show a similar behavior, where the fundamental frequency from 12 Hz for $P_1$ position decrease to 10 Hz in position $P_2$ (figure 9) and 8 Hz for $P_3$ position (figure 10).

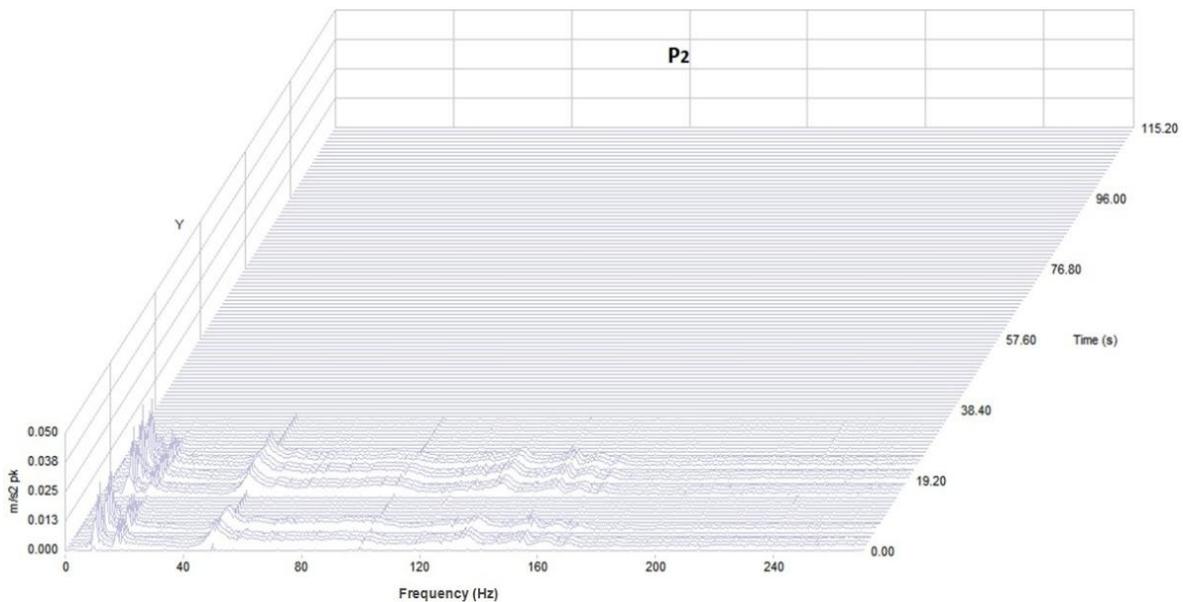

Figure 9: Waterfall frequency diagram for Y direction in $P_2$ position

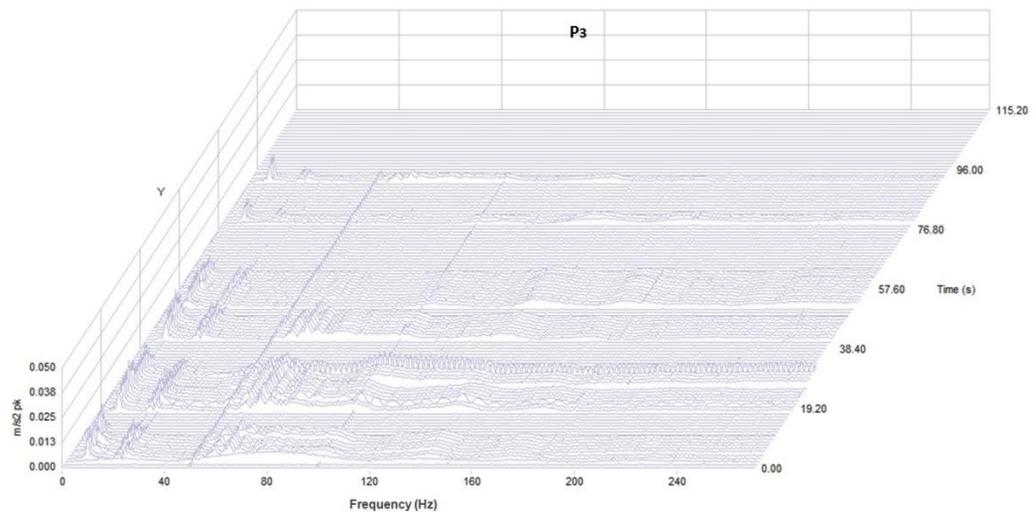

Figure 10: Waterfall frequency diagram for Y direction in $P_3$ position

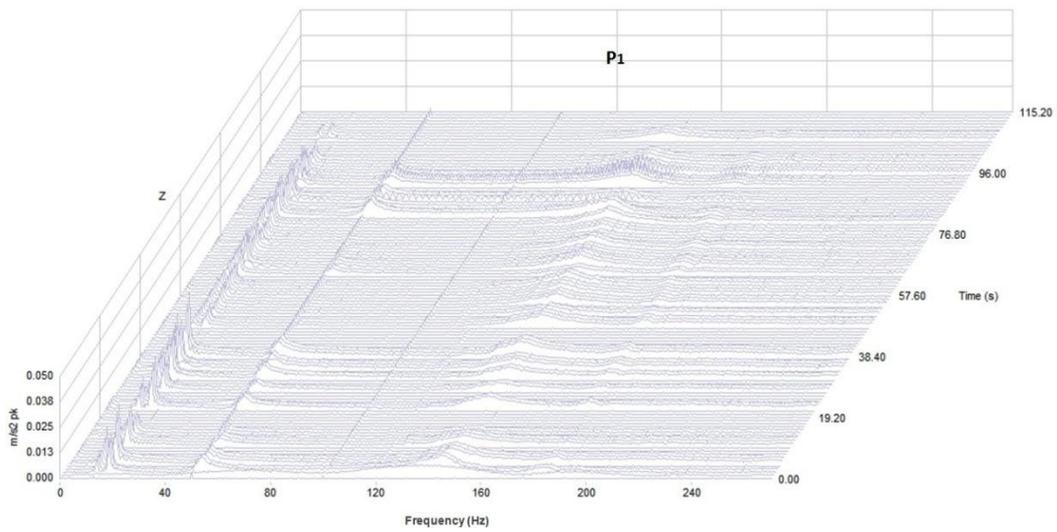

Figure 11: Waterfall frequency diagram for Z direction in $P_1$ position

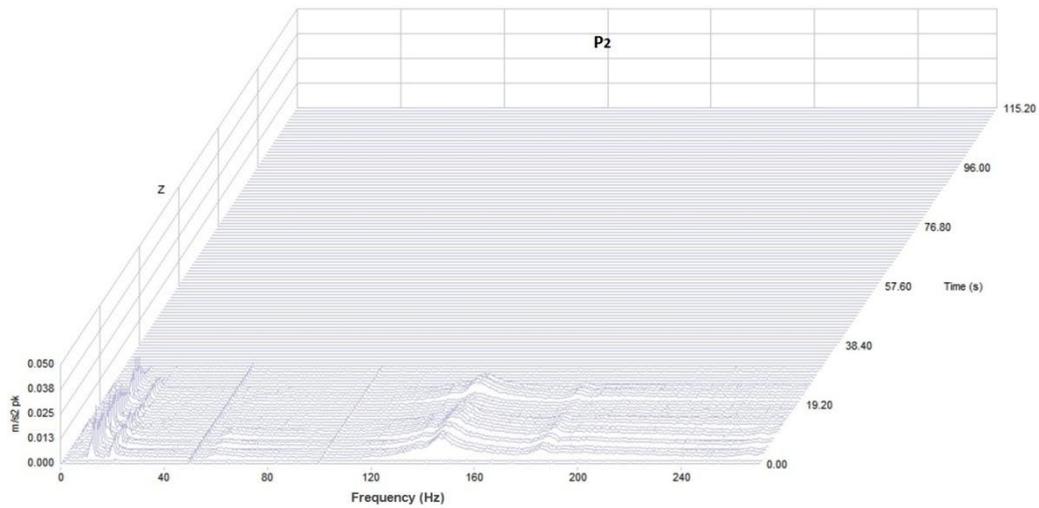

Figure 12: Waterfall frequency diagram for Z direction in $P_2$ position

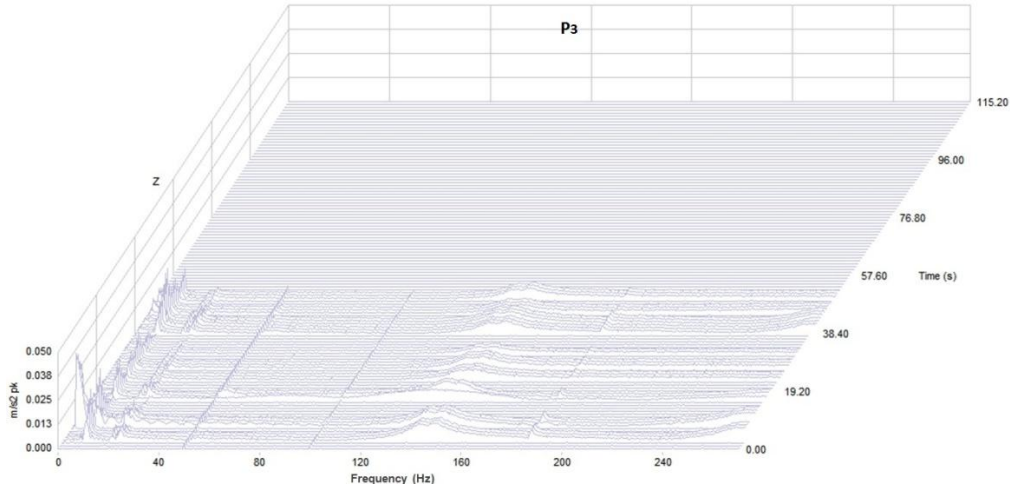
Figure 13: Waterfall frequency diagram for Z direction in $P_3$ position

The self excited frequencies in the Z direction (figure 1) show the same trend as the Y direction. The robot arm rigidity decreases, from position $P_1$ (figure 11) in relation with $P_2$ position (figure 12) and the $P_3$ position (figure 13). After analyzing the self-excited frequencies measured from the impact test to see a sensible increase in stiffness in the X direction and a decrease in stiffness in the direction Y and Z. The variation of the low frequency is about 10% at position $P_1$ to position $P_2$ respectively $P_3$ in the three directions.

The analysis of the static frequency aims to impact frequency location of the robot in different configurations and their variation according to different position on the robot to perform. Thus obtain an overview of the self-excited frequency, where there is slight increase frequency in the X direction and a very small decrease of frequency in the Y and Z direction at $P_1$ position from $P_2$ and $P_3$ position.

## 4 Vibrations analysis during the spindle rotation

Frequency analysis is performed in the dynamic case for the rotation speed of 12,032 rpm which can be identified both LFR and HFR frequency range, figure 12 in the $P_1$ position. The choice of tool speed is considering cutting parameters for testing and also the comparative material will be made between the behavior of the robot during the cutting process and outside the cutting process. The range of the high frequency represents the excitation of the component elements of the robot arm, representing one of the prospects of this research [10], [11]. By the spectrum analysis synchronized with rotational speed can identify in the case of LRF the existence an imbalance revealed by the first order of a harmonic frequency and also a misalignment corresponding with the second order of harmonic and. The frequency 480 Hz represents the critical frequency because it is a self-excited frequency measured in $P_1$ position (figure 14). Vibration analysis during the rotation speed (without cutting) is absolutely necessary for the excitation of self-excited frequencies of the robot but also to identify certain defects, such as imbalance, misalignment, faulty grip, bearing failure, the variation of electrical parameters, etc. Taking into account the initial phase of the robot a detailed dynamic analysis components is viable both for the quality of work of the robot and to develop a maintenance plan.

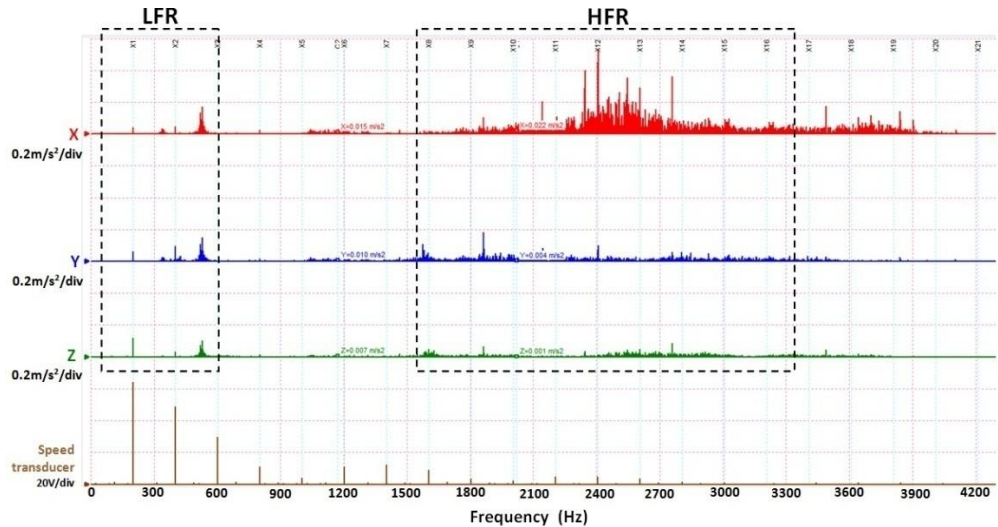

Figure 14: Frequency spectrum on x, y and z direction without cutting

## 5 Vibrations analysis during the milling process

Milling tests are performed on a resin type material loaded with Aluminum with a feed rate of 1250 mm/min. A milling tool with 6 mm diameter with 6 teeth is used at the cutting speed of 227 m/min. Dynamic analysis of the robot is performed during processing which highlight vibration amplitude generated by the milling process in two configurations: the position $P_1$ and $P_2$ position. Note the important difference amplitudes in the two situations. Since the X direction increases rigidity in position $P_1$ and $P_2$ to the position processing occurs in the same direction of the material, we can see that the amplitude in the direction X in position $P_1$ are much higher than in the $P_2$ position, also the amplitudes of the Y direction and Z are higher in position $P_1$ from position $P_2$.

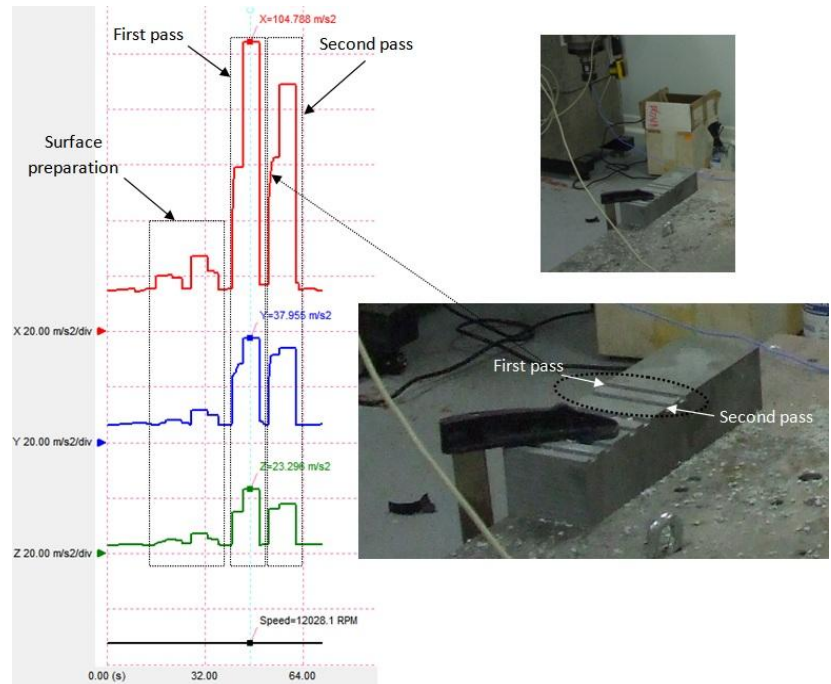

Figure 15: The trend signal on X, Y and Z directions during the milling process in $P_1$ position

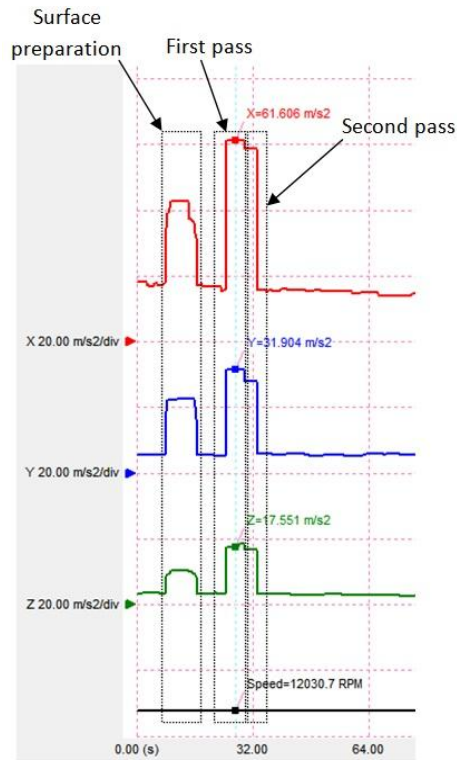

Figure 16: The trend signal on X, Y and Z direction during the milling process in $P_2$ position

The objective of this work is to characterize the dynamic behavior of the robot's ability to work. For a better understanding of dynamic movement during processing to perform a waveform signal and frequency analysis of the two working positions $P_1$ and $P_2$ (figure 17 and figure 18). Impact measured at lower frequencies (LFR) is not in the spectrum and therefore we apply the envelope method in order to verify the existence of lower frequencies measured. The vibration amplitudes in $P_1$ position are higher from $P_2$, which shows better stiffness in $P_2$ position from $P_1$ position.

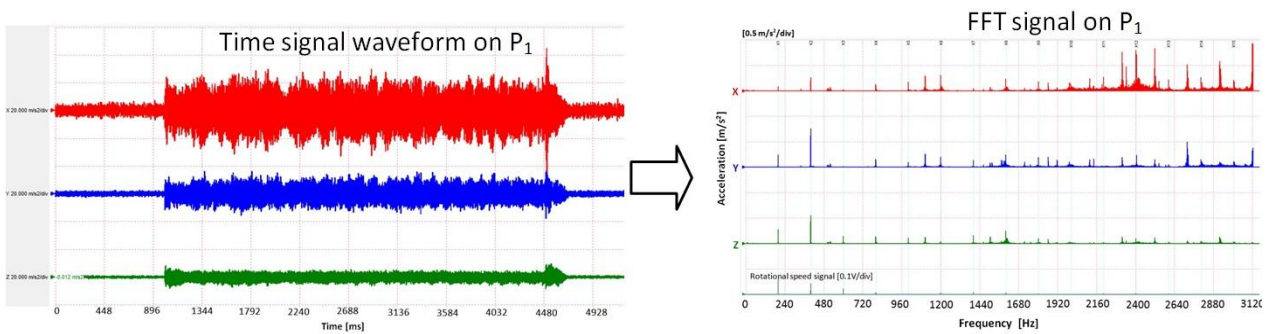

Figure 17: The waveform signal and the spectrum frequency on X, Y, Z direction during the milling process in $P_1$ position

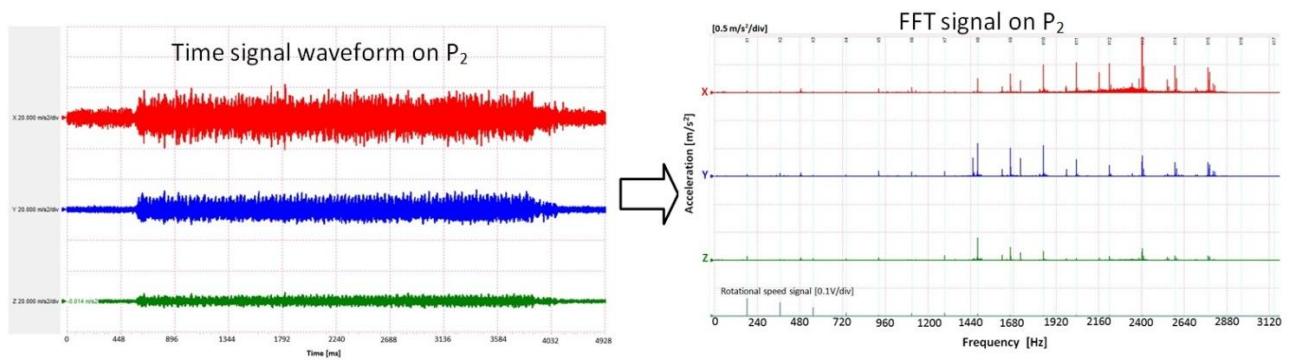

Figure 18: The waveform signal and the spectrum frequency on X, Y, Z direction during the milling process in $P_2$ position

The aim of the envelope method applied to the robot behaviour during the milling process is achieved by frequency domain processing, consistent in high accuracy synchronous FFT transform, filtering resonance band of workpiece and tool, Hilbert transform [12], [13], [14] followed by Inverse Fast Fourier Transform (IFFT). Next the FFT analyses of the envelope ensure high precision description of the milling tool to identify the type and amplitude of asymmetry and wear. Each cutter tooth asymmetry is automatically qualified through the harmonic components with a lower frequency than the principal frequency equivalent of teeth number. To detect structural defects that may occur in these machine components, spectral analysis of the signal's envelope has been widely employed [13], [14]. This is based on the consideration that structural impacts induced by a localized defect often excite one or more resonance modes of the structure and generate impulsive vibrations in a repetitive and periodic way. Frequencies related to such resonance modes are often located in higher frequency regions than those caused by the robot-borne vibrations, and are characterized by an energy concentration within a relatively narrow band centred at one of the harmonics of the resonance frequency. By utilizing the effect of mechanical amplification provided by structural resonances, defect-induced vibration features can be separated from the background noise and interference for diagnosis purpose [15]. The resonance filter band for envelope method is between 2000 - 3000Hz.

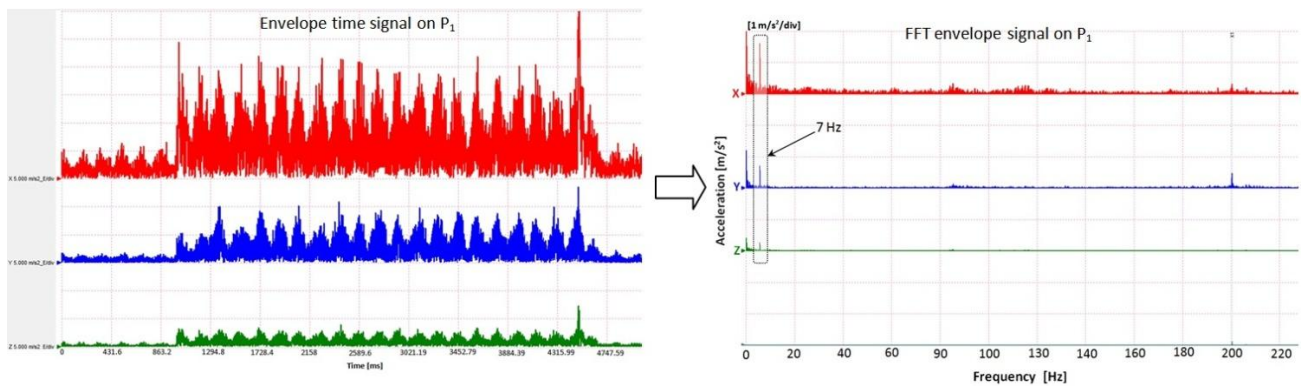

Figure 19: The envelope time signal and the spectrum frequency envelope on X, Y, Z direction during the milling process in $P_1$ position

After the application of envelope is emphasized amplitude low frequency range contained in 7 - 11 Hz that is generated by varying cutting forces resulting in excitation resonance frequency of the robot. High-frequency amplitude modulation generated by the contact tool/chip/workpiece allows the Hilbert transform [16] their impact frequency identification robot. An important perspective is to locate the source of these lower frequency vibrations. This robot can validate that increases rigidity to the $P_1$ position from position $P_2$.

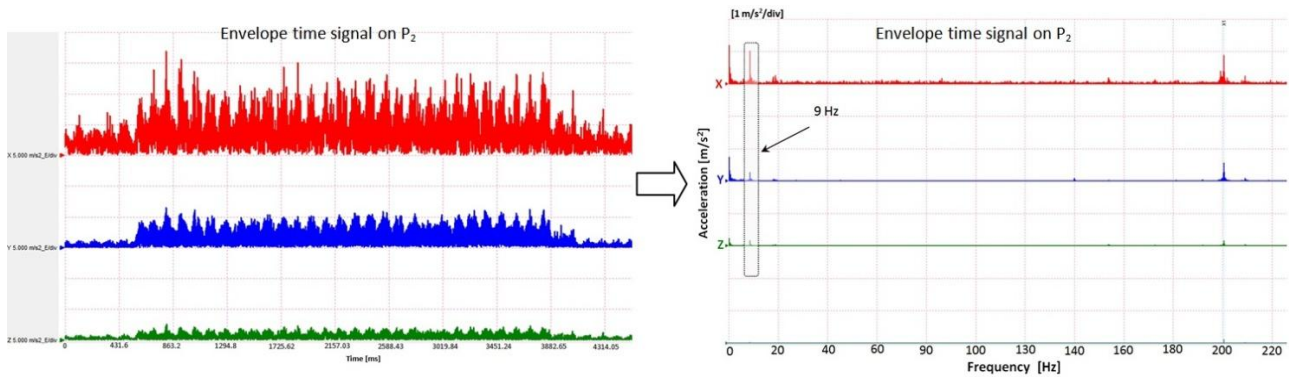

Figure 20: The envelope time signal and the spectrum frequency envelope on X, Y, Z direction during the milling process in $P_2$ position

# 6 Conclusions

Robots are increasingly used in machining applications where the performance of position and orientation of the tool center point is important. The use of robots in machining applications takes a large scale, in particular in the aerospace industry. Knowing that the dynamic behavior of robots used in processing materials is poor in comparison with a CNC gantry, a detailed approach is required concerning the dynamical behavior induced by the cutting process. This paper aims at characterizing the dynamics change of the robot dynamical behavior through several points of the workspace. Especially the evolution of its stiffness for different configurations of the workspace is analyzed. An experimental protocol was designed and developed to highlight the dynamic characteristics of the robot This analysis is done in three stages: The first step is consisting on determining the self-excited frequencies in different configurations of work. The second phase aims to analyze dynamic vibration in the same time with the rotational speed. Finally, the third stage consists of vibration analysis during the milling process. Frequency analysis showed a small change in owns way their three configurations depending on the position. Measured frequencies were divided into two categories-Low Frequency Range LFR and HFR-High Frequency Range, putting it out through LFR frequency Waterfall diagram. The robot's frequencies are identified in the case while the dynamic cutting process has been a significant increase in vibration basic configuration close to the robot base position $P_1$ from position $P_2$. This increase in vibration is based on the fact that stiffness in the direction X is greater in position $P_2$ to the $P_1$ position.

Dynamic analysis of robot behavior during the cutting process is performed through to enveloping method for low frequencies resonance identification, located in the range of 7-11 Hz, associated of the robot components. In perspective, an analysis of cutting forces measured during milling process correlated with measured vibrations will be made in order to highlight the dynamic behavior in different work configurations as well as analysis of surface regeneration. Spectrum cartography and a map of the stiffness of the robot workspace is one of the main insights of this research.

On the other hand, further work plane at the creation of a model to optimize cutting parameters in order to obtain stability during the cutting process. Frequency results are needed both to highlight critical frequency and supply the dynamic model for the optimization and the monitoring of robotics machining tasks.

**Acknowledgement**: This paper was supported by CNCSIS-UEFISCSU, project PNII-RUcode194/2010.

# References

[1] M. Summer, Robot capability test and development of industrial robot positioning system for the aerospace industry. In: SAE 2005 AeroTech Congress & Exhibition, Grapevine, TX, SAE Technical Papers 2005-01-3336, 2005.

[2] W. Khalil, E. Dombre, Modeling, Identification and Control of Robots, Hermes Science Publications, 2002.

[3] C. Lecerf-Dumas, S. Caro, M. Cherif, S. Garnier, B. Furet, A New Methodology for Joints Stiffness Identification of Serial Robots, IEEE/RSJ International Conference on Intelligent Robots and Systems (IROS 2010) Taipei Editorial System, 2010.

[4] J. Angeles, Fundamentals of Robotic Mechanical Systems, Theory, Methods and Algorithms, Third Edition, Springer, New York, 2007. (First Edition published in 1997).

[5] P. Bourdet, L. Mathieu, C. Lartigue, A. Ballu, The concept of the small displacement torsor in metrology, Series on Advances in Mathematics for Applied Sciences, Advanced mathematical tools in metrology II, 40, pp. 110-122, World Scientific, 1996.

[6] F. Girardin, Etude de l'usinage de matériaux performants et surveillance de l'usinage, L'Institut National de Sciences Appliquées de Lyon, 2010.

[7] D. A. Axinte, N. Gindy, K. Fox, I. Unanue, "Process monitoring to assist the workpiece surface quality in machining", International Journal of Machine Tools & Manufacture, 44, pp. 1091–1108, 2004.

[8] Liang, J. and Bi, S. Design and experimental study of an end effector for robotic drilling, International Journal of Advanced Manufacturing Technology, 50, pp. 399-407, 2010, doi :10.1007/s00170-009-2494-9.

[9] C.F. Bisu, J-Y K'nevez, P. Darnis, R. Laheurte, A. Gérard, "New method to characterize a machining system: application in turning" International Journal of Material Forming. 2, (2), pp. 93-105, 2009, DOI : 10.1007/s12289-009-0395-y,

[10] C. F.Bisu, M. Zapciu, A. Gérard, V. Vijelea, M. Anica, "New approach of envelope dynamic analysis for milling process", Eighth International Conference on High Speed Machining, Metz, France, Dec. 8-10, 2010.

[11] Shin-Ichi Metsouoka, Kazunori Schimizu, Nobuyuki Yamazaki, Yashunori Oki, High speed end milling of an articulated robot and its characteristics, Journal of Material Processing Technology, 95, (1-3), pp. 83-89, 1999.

[12] I. Zaghbany, V. Songmene, Estimation of machine-tool dynamic parameters during machining operation through operational modal analysis, International Journal of Machine Tools & Manufacture, 49, (12-13), pp. 947–957, 2009.

[13] R. Yan, R. X. Gao, "Multi-scale enveloping spectrogram for vibration analysis in bearing defect diagnosis", Tribology International, 42, pp. 293–302, 2009.

[14] T.Kalvoda; Y.R. Hwang; "A cutter tool monitoring in machining process using Hilbert-Huang transform"; In: International Journal of Machine Tools and Manufacture, 50, (5), pp. 495-501, 2010.

[15] V. Gagnol, T-P. Le, P. Ray, "Modal identification of spindle-tool unit in high-speed machining", Mechanical Systems and Signal Processing, 25, (11), pp. 2388-2398, 2011, doi:10.1016/j.ymssp.2011.02.019.

[16] X. Wang, "Numerical Implementation of the Hilbert Transform", Thesis, University of Saskatchewan, Saskatoon, 2006.